\documentclass[letterpaper, 10 pt, conference]{ieeeconf}

\usepackage{amsmath, amssymb, mathtools}
\usepackage{svg}
\usepackage{subcaption}
\newcommand\blockkron{\stackrel{\mathclap{\normalfont\mbox{b}}}{\otimes}}

\IEEEoverridecommandlockouts
\title{\LARGE \bf
Space-Time Continuum: Continuous Shape and Time State Estimation for Flexible Robots
}

\author{Spencer Teetaert$^{1,2}$, Sven Lilge$^{1}$, Jessica Burgner-Kahrs$^{2}$, and Timothy D. Barfoot$^{1}$
\thanks{$^{1}$Autonomous Space Robotics Laboratory, University of Toronto, Canada. {\tt\small spencer.teetaert@robotics.utias.utoronto.ca}}%
\thanks{$^{2}$Continuum Robotics Laboratory, University of Toronto, Canada.}%
}

\begin{document}

\maketitle
\thispagestyle{empty}
\pagestyle{empty}

\section{INTRODUCTION}
\begin{figure}
   \centering
   \def\svgwidth{0.48\textwidth}
\begingroup%
  \makeatletter%
  \providecommand\color[2][]{%
    \errmessage{(Inkscape) Color is used for the text in Inkscape, but the package 'color.sty' is not loaded}%
    \renewcommand\color[2][]{}%
  }%
  \providecommand\transparent[1]{%
    \errmessage{(Inkscape) Transparency is used (non-zero) for the text in Inkscape, but the package 'transparent.sty' is not loaded}%
    \renewcommand\transparent[1]{}%
  }%
  \providecommand\rotatebox[2]{#2}%
  \newcommand*\fsize{\dimexpr\f@size pt\relax}%
  \newcommand*\lineheight[1]{\fontsize{\fsize}{#1\fsize}\selectfont}%
  \ifx\svgwidth\undefined%
    \setlength{\unitlength}{2430bp}%
    \ifx\svgscale\undefined%
      \relax%
    \else%
      \setlength{\unitlength}{\unitlength * \real{\svgscale}}%
    \fi%
  \else%
    \setlength{\unitlength}{\svgwidth}%
  \fi%
  \global\let\svgwidth\undefined%
  \global\let\svgscale\undefined%
  \makeatother%
  \begin{picture}(1,0.59259259)%
    \lineheight{1}%
    \setlength\tabcolsep{0pt}%
    \put(0,0){\includegraphics[width=\unitlength,page=1]{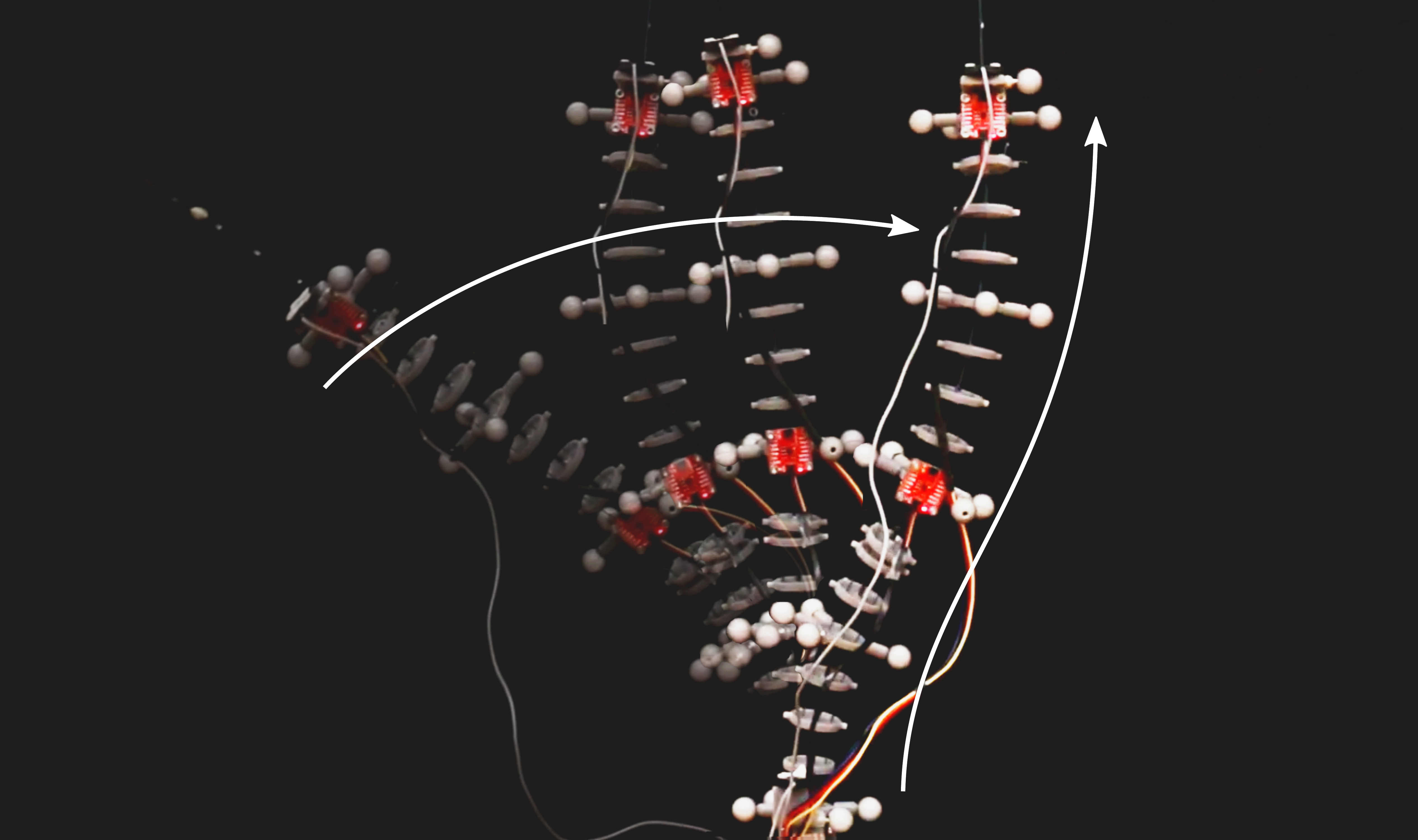}}%
    \put(0.79259259,0.3962963){\color[rgb]{1,1,1}\makebox(0,0)[lt]{\lineheight{1.25}\smash{\begin{tabular}[t]{l}Continuous \\Arclength\end{tabular}}}}%
    \put(0.17654321,0.42839506){\color[rgb]{1,1,1}\makebox(0,0)[lt]{\lineheight{1.25}\smash{\begin{tabular}[t]{l}Continuous \\Time\end{tabular}}}}%
  \end{picture}%
\endgroup%

   \caption{ A continuum robot is shown at multiple timesteps with increasing opacity. The prototype has four types of asynchronous sensors: gyroscopes, 6 DOF pose sensors, a strain sensor, and an external motion capture. }
   \label{fig:robot_prototype}
\end{figure}

State estimation for continuum robotics is a comparatively underdeveloped area. Many approaches make use of the quasi-static assumption \cite{Mahoney2016a,Lilge2022}. This approach reduces model complexity, negating the need for dynamic modelling, resulting in generally faster computation speeds. Several dynamic estimators have been proposed. In \cite{Lobaton2013,Ataka2016} the authors make use of Kalman filtering techniques. A typical assumption these dynamic estimators make is that of a constant curvature shape, further limiting the accuracy of the method in pursuit of model simplicity. These methods discretize the estimators in time. In mobile robotics, \cite{Anderson2015} introduces the use of Gaussian-process (GP) regression for performing continuous-time state estimation of a mobile robot. The authors highlight the motivation for continuous-time estimation, mainly its ability to better handle scanning-while-moving sensors. In \cite{Lilge2022} this GP method is co-opted for use in continuous-space estimation of a continuum robot at a single point in time. Continuous representations for such robots provide representation over the full robot shape, granting more complex reasoning capability further down the robotics pipeline (such as in control and planning). Motivated by the same continuous-time reasoning from \cite{Anderson2015} applied to continuum robots, we move towards a more principled way of handling asynchronous measurements while estimating continuum robot shapes. A robot that would benefit from our work is that of Fig. \ref{fig:robot_prototype}, where multiple sensors produce data at varying frequencies. This scenario is growing more common in continuum robotics applications where a variety of dense sensor information is more readily available. This paper has two main contributions: First, we propose a two-dimensional GP prior that combines the benefits of \cite{Lilge2022} and \cite{Anderson2015} and has a sparse block-tridiagonal inverse kernel matrix, enabling efficient solving speeds. By regressing a 2D GP, we achieve smoothness of our full state over both spatial and temporal dimensions, a feat that was previously not achievable using quasi-static or spatially discretized estimators. Second, we demonstrate the resulting space-time estimator on a simulated continuum robot. To the best of our knowledge, this is the first state estimator proposed for continuum robots that is continuous in both space {\em and} time.

\section{SPACE-TIME PRIOR}

We describe the continuum robot state over space and time, extending the formulations proposed in \cite{Anderson2015} and \cite{Lilge2022}, using the expression: 
\begin{align}
   \label{eq:prior}
   \nonumber \boldsymbol{x} & (s, t) = \boldsymbol{\Phi}(s,s_0;t,t_0)\boldsymbol{x}(s_0,t_0)                                                                                                          \\
   \nonumber                & + \int_{s_0}^{s}\boldsymbol{\Phi}(s,\sigma;t,t_0)(\boldsymbol{v}_s(\sigma) + \boldsymbol{L}_s\boldsymbol{w}_s(\sigma))d\sigma                                           \\
   \nonumber                & + \int_{t_0}^{t}\boldsymbol{\Phi}(s,s_0;t,\tau)(\boldsymbol{v}_t(\tau) + \boldsymbol{L}_t\boldsymbol{w}_t(\tau))d\tau                                                   \\
                            & + \int_{t_0}^{t}\int_{s_0}^{s}\boldsymbol{\Phi}(s,\sigma;t,\tau)(\boldsymbol{v}_{st}(\sigma, \tau) + \boldsymbol{L}_{st}\boldsymbol{w}_{st}(\sigma, \tau))d\sigma d\tau,
\end{align}
where $\boldsymbol{x}(s, t)$ is a state at time $t$ and arclength $s$, $\boldsymbol{\Phi}(s,s';t,t')$ is a transition function from a state at $(s,t)$ to a state at $(s',t')$, $\boldsymbol{v}(\cdot)$ are input functions, $\boldsymbol{L}$ are selection matrices, and $\boldsymbol{w}(\cdot)$ are Gaussian noise terms.
The mean and covariance of this function act as our GP prior.
Armed with this 2D prior, we continue by stacking our state and covariance values \cite{Barfoot2017}. We first stack over the spatial dimension followed by the temporal one. This ultimately leads us to the following representation for an inverse kernel matrix:

\begin{align}
   \nonumber \check{\boldsymbol{P}}^{-1} & = \boldsymbol{A}^{-T} \boldsymbol{Q}^{-1} \boldsymbol{A}^{-1}                                                                       \\
                                         & = (\boldsymbol{A}_t^{-T} \blockkron \boldsymbol{A}_s^{-T}) \boldsymbol{Q}^{-1} (\boldsymbol{A}_t^{-1} \blockkron \boldsymbol{A}_s^{-1}),
\end{align}
where $\blockkron$ denotes the block Kronecker product \cite{kishka18}, $\boldsymbol{A}_t$ and $\boldsymbol{A}_s$ denote the lifted matrices from the systems in \cite{Anderson2015} and \cite{Lilge2022} defined using the method from \cite{Barfoot2017}, and $\boldsymbol{Q}$ is a stacked matrix of the prior covariance estimates at each node in our factor graph. 

\begin{figure}[h]
   \centering
   \def\svgwidth{0.45\textwidth}
   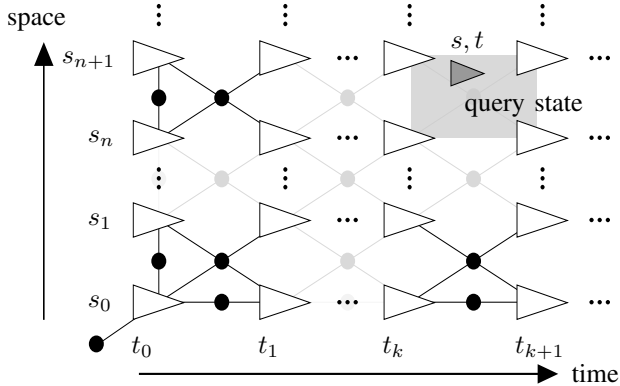
   \caption{ Factor graph for space-time estimation of a continuum robot. With the GP formulation, continuous querying in both arclength and time is available in $O(1)$ time. }
   \label{fig:lattice}
\end{figure}

This inverse kernel matrix enjoys a depth-two recursive block-tridiagonal structure, a fact we can take advantage of for improved runtime. Considering $N$ spatial and $K$ temporal nodes, this system can be solved in $O(KN)$ time. The corresponding factor graph for this inverse kernel matrix is presented in Fig. \ref{fig:lattice}. The graph contains binary factors similar to those from \cite{Anderson2015} and \cite{Lilge2022}, a unary factor linked to the initial condition, and the addition of new quaternary factors each linking four nodes in the graph.

\begin{figure}[h]
   \centering
   \includegraphics[width=0.91\linewidth]{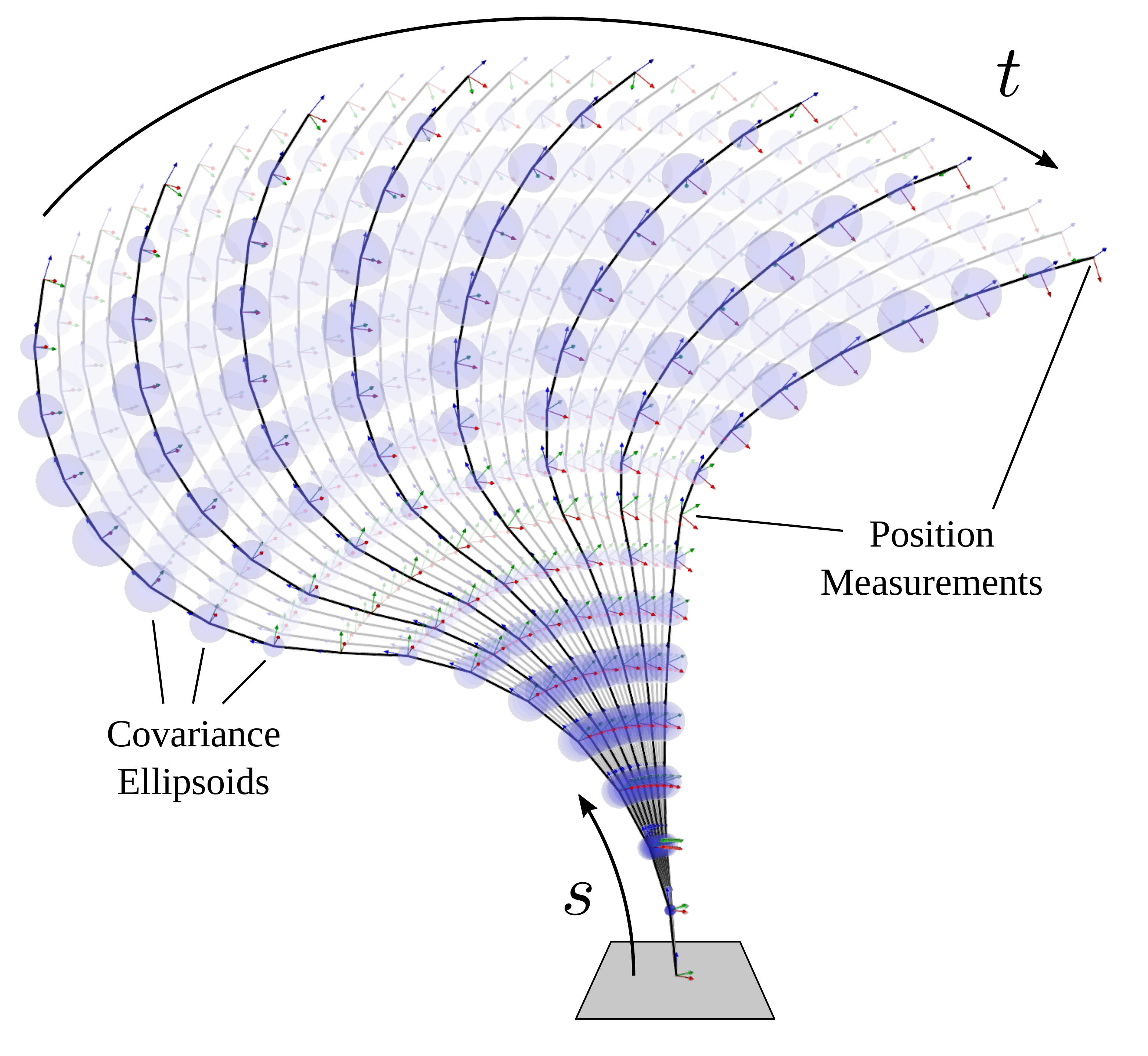}
   \caption{Example state estimation result of the proposed approach using a simulated tendon-driven continuum robot. Sensing includes noisy strain measurements at every estimation node as well as two noisy discrete position measurements. The state estimate includes the smooth pose and strain of the continuum robot over both space (arclength) $s$ and time $t$. The state mean is shown in black and via coloured coordinate frames. The state uncertainty is visualized using blue covariance ellipsoids.}
   \label{fig:estimator}
\end{figure}

\section{APPLICATION}

We use the presented 2D prior and new quaternary error terms in a continuous-space-and-time estimator for continuum robots. We estimate the pose, strain, velocity, and strain-velocity of the robot at each node in our factor graph over space and time. We represent the pose component of our state as elements of $SE(3)$, allowing us to solve the batch problem through unconstrained non-linear optimization. We linearize the error terms present in our factor graph and iterate over the entire batch solution. We employ a sparse Cholesky solver to take advantage of the system's structure. As we formulated our estimator as a GP, we also gained GP interpolation capability. This is what allows us to query our estimator continuously in both arclength and time.
See Fig.~\ref{fig:estimator} for an example result of the proposed estimator.

\section{DISCUSSION AND FUTURE WORK}

The proposed space-time estimator offers a principled way to jointly estimate the continuous shape and velocity of a continuum robot. By formulating the estimation problem as a factor graph optimization, we leave open the possibility for the inclusion of other well-established pose-graph features from other research areas (such as SLAM, numerous measurement models, joint robot configurations, etc.). These have historically been left out entirely or required great difficulty to integrate with continuum robot estimation methods.

This work may be applied to other applications that are parameterized by two independent variables. Examples include quasi-static estimation of a surface and shape estimation of serpentine robots. In future publications we hope to expand on the theory of this work, providing derivations for the properties claimed here (such as continuous interpolation), as well as demonstrate its usefulness on physical robots. 

\bibliographystyle{IEEEtran}
\bibliography{IEEEabrv,refs}

\end{document}